\documentclass[9pt,twocolumn,twoside]{pnas-new}

\templatetype{pnasresearcharticle} 

\title{Learning offline: memory replay in biological and artificial reinforcement learning}

\author[1,*]{Emma L. Roscow}
\author[2]{Raymond Chua} 
\author[3]{Rui Ponte Costa}
\author[4,6]{Matt W. Jones}
\author[5,6]{Nathan Lepora}

\affil[1]{Centre de Recerca Matemàtica, Bellaterra, Spain}
\affil[2]{McGill University and Mila, Montréal, Canada}
\affil[3]{Bristol Computational Neuroscience Unit, Intelligent Systems Lab, Department of Computer Science, University of Bristol, UK}
\affil[4]{School of Physiology, Pharmacology and Neuroscience, University of Bristol, Bristol, UK}
\affil[5]{Department of Engineering Mathematics and Bristol Robotics Laboratory, University of Bristol, Bristol, UK}
\affil[*]{Correspondence: eroscow@crm.cat}
\affil[6]{These authors contributed equally to this work}

\leadauthor{Roscow}


\begin{abstract}
Learning to act in an environment to maximise rewards is among the brain’s key functions. This process has
often been conceptualised within the framework of reinforcement learning, which has also gained prominence in machine learning and artificial intelligence (AI) as a way to optimise decision‐ making. A common aspect of both biological and machine reinforcement learning is the reactivation of previously experienced episodes, referred to as replay. Replay is important for memory consolidation in biological neural networks, and is key to stabilising learning in deep neural networks. Here, we review recent developments concerning the functional roles of replay in the fields of neuroscience and AI. Complementary progress suggests how replay might support learning processes, including generalisation and continual learning, affording opportunities to transfer knowledge across the two fields to advance the understanding of biological and artificial learning and memory.
\end{abstract}

\begin{document}

\maketitle
\thispagestyle{firststyle}
\ifthenelse{\boolean{shortarticle}}{\ifthenelse{\boolean{singlecolumn}}{\abscontentformatted}{\abscontent}}{}

\section*{Replay in biological and artificial reinforcement learning}
Research into \textbf{reinforcement learning} (see Glossary) in biology, psychology, and AI has a long
and symbiotic history [1]. In recent years, \textbf{deep reinforcement learning} has shown remarkable success in problems previously thought intractable. Key to the success of these algorithms is the practice of interleaving new trials with old ones, a technique known as \textbf{experience replay} [2], and an example of convergence between biological and artificial neural networks. Replay has proved important for cognitive theories of memory [3], and the importance of replay in deep reinforcement learning supports the theory that ‘offline’ activity can aid learning and memory, raising general computational principles through which this can be achieved by any intelligent system. The relative ease of manipulating replay in artificial systems compared with biological brains means that advances in deep reinforcement learning can offer useful test cases for neuroscientists. Meanwhile, the continuing search for ever-better learning algorithms in AI could be informed by recent neurobiological insights into how self-organised activity can lead to selective strengthening or weakening of specific memories [4–8], generalisation from individual experiences to abstract knowledge [9], and, more generally, flexible adaptation to a
changing world [10–13].

Here, we review recent developments in both fields. We discuss the successes of uniform experience replay and prioritised experience replay in AI, reviewing analogous neurobiological phenomena evident in rodent and human experiments. Our review converges on the proposal that methods which avoid explicit storage of past trials and generate their own replay samples may better reflect biological processes and hold the key to flexible reinforcement learning.

\afterpage{\begin{mdframed}[backgroundcolor=black!10,rightline=false,leftline=false,topline=false,bottomline=false,innerbottommargin=16pt,innerrightmargin=20pt,afterbreak={\hfill}]
\section*{Highlights}
\begin{itemize}
  \item Reinforcement learning in deep neural networks often relies on the interleaving of new and old episodes, a technique which mimics the replay of neuronal activity in the brain.

    \item Biological replay is important for memory consolidation, and has wider roles in other cognitive processes such as planning and generalisation. Deep neural networks offer a framework for understanding the role of replay in learning and memory.

    \item We suggest how recent developments could be leveraged to support more robust, efficient, and flexible reinforcement learning agents by avoiding explicit storage of past trials.

    \item Theoretical advances and more sophisticated experimental task designs will help uncover how biological replay supports complex cognition over time and throughout the brain.
\end{itemize}

\end{mdframed}}

\section*{Uniform experience replay and its origins in neuroscience}
Reinforcement learning in AI was developed in the mid-20th century, taking inspiration from earlier animal behaviour research [1]. In training reinforcement learning algorithms, an artificial agent collects data samples through continuous interaction with a real or simulated world, learning policies for selecting actions given the state of the environment in a way that maximises a reward function. Given limited online experience, learning can be accelerated by storing past experiences and subsequently sampling from them repeatedly, in effect to increase the training set.

Experience replay first appeared in the AI literature in the early 1990s as a means to achieve such an increase [2] (Box 1), and grew in popularity with the advent of deep reinforcement learning and its applications to Atari games and Go in the early 2010s [14,15]. Independently, a series of neurophysiology studies beginning in the 1980s and 1990s found a similar phenomenon of reusing past experience in the mammalian brain (see Figure 1, Box 2 and recent reviews for more detail [3,16–19]). These neuroscientific replay studies unveiled, among other insights, potential mechanisms of sleep-dependent \textbf{memory consolidation}, with replay providing a cellular basis for the long-standing observation that sleep supports memory [20].

How does replay improve memory? One of the leading hypotheses is that replay induces Hebbian \textbf{plasticity} between the cells being replayed [21–25], thereby strengthening their synaptic connections. Replay events, particularly during non-rapid‐eye‐movement (\textbf{non-REM}) \textbf{sleep}, typically reiterate neural patterns on a faster timescale than during the original experience [26–28], which might further encourage spike-timing-dependent plasticity [21]; one could call this the ‘offloading plasticity until later’ theory of how replay supports memory consolidation. However, while the importance of replay for spatial (hippocampus-dependent) memory has been established, questions remain about which aspects of the experience are represented in replayed activity, how replay patterns propagate through the brain, and the
roles of replay in wider cognitive processes.

Computational studies have suggested functions for replay that extend beyond memory consolidation (Box 3). The complementary learning systems theory has used the so-called ‘penguin problem’ to illustrate the necessity of maintaining a network that is stable enough for acquired knowledge to persist, but plastic enough to incorporate new knowledge [29]. In this illustration, a network was trained to classify living things from their characteristics, before being presented with an anomalous semi-aquatic bird (the penguin) that has feathers and wings like a bird, but does not fly and swims like a fish. Updating the connection weights to incorporate this new information disrupted and worsened performance for other birds [29]. The proposed solution was replay, interleaving training of the new item (penguin) with older, similar items (other birds); this proved sufficient to maintain both representations without interference. One could call this the ‘preventing \textbf{catastrophic interference}’ theory of how replay improves memory consolidation, an idea that dates back to connectionist models early in the history of
artificial neural networks [30].

\afterpage{\begin{mdframed}[backgroundcolor=black!10,rightline=false,leftline=false,topline=false,bottomline=false, innerbottommargin=12pt]
\section*{Box 1.  Experience replay in artificial intelligence}
In discrete time sequences, incoming data samples are usually represented in the form of an experience tuple, consisting of a state s at time step t, action a performed at time step t, reward r obtained at time step t, and the next state st+1 at next time step t + 1. This \textbf{experience tuple} is first stored in a buffer and, during the learning phase, samples are randomly drawn in mini-batch from this buffer uniformly.

In deep Q networks, these mini-batch samples are then used to learn the agent’s Q-value function, the expected future reward associated with each pair of states and actions, using off-policy Q-learning. The Q-value function is policy-dependent as it relies on data collected resulting from the agent’s actions derived from its policy (behaviour). In the tabular setting, this Q-value function can be represented by a table of size $\vert S\vert \times \vert A\vert$, where $\vert S\vert$ is the number of states and $\vert A\vert$ is the number of actions in the environment. It is defined as:

$$ Q^ \pi (s, a) = E_\pi [r_1 + \gamma r_2 + \gamma^2 r_3 + ... \vert s_0 = s, a_0 = a] $$

where $\gamma$ is the discount factor that controls how much the agent prioritises immediate rewards against long-term rewards. The off-policy update rule for the Q-value function is:

$$ Q(s, a) \leftarrow Q(s, a) + \alpha [r_{t+1} + \gamma \max\limits_{a'} Q(s', a')] $$

where $\alpha$ is the learning rate. The Q-value function is said to have been completely learnt when its values have converged.

As the agent continuously explores the environment and collects data samples, sooner or later the buffer will become full, and the oldest samples will be replaced by newer samples. This strikes a balance between learning the most recent samples and allowing older samples to ‘live’ longer than they usually would, such as in the classical online learning setting. Experience replay has shown to improve the learning efficiency of artificial agents [105].

\end{mdframed}}

Models of replay are often concerned with its role in reinforcement learning. Although biological replay is discussed commonly in terms of episodic memory consolidation and the integration of new memory traces into long-term storage [3,16–19,29,31–37], evidence from animal studies usually comes from spatial navigation tasks in which food rewards or electrical brain stimulation are used to reinforce exploration and navigation of an environment [4,5,8,10–12,26,27,29,31–48]. The additional plasticity that replay incurs may itself reinforce habitual behaviours that are driven by the replayed activity patterns [39]. Replay of activity in the hippocampus alone is necessary for stabilising newly formed representations of the environment, ensuring that learned \textbf{state transitions} are maintained for subsequent visits [49,50]. In addition, the recruitment of other brain areas that are involved in evaluating likely outcomes and rewards during replay may promote further updating of stored action values or state values in neural reinforcement learning circuits [13,40,48,51].

These two functional roles of replay (preventing catastrophic interference and facilitating reinforcement learning) are particularly relevant for deep reinforcement learning. The fact that the use of experience replay was crucial to the first notable success of reinforcement learning with deep neural networks showed that these proposed functions of replay have application beyond the mammalian brain and extend to artificially intelligent systems [14]. In this work, an artificial neural network composed of several convolutional and fully connected layers received visual input (images from Atari 2600 computer games) while producing joystick movements to play the game. Its learning algorithm builds on classical \textbf{Q-learning}, which maps states (visual input) to actions (joystick output). The error generated by the Q-learning loss function is then used to train the deep neural network using the \textbf{backpropagation} algorithm. This process gradually optimises the neuron parameters (e.g., synaptic weights) towards an optimal mapping between the state space (Atari images) and the possible actions (Figure 2).

\afterpage{\begin{mdframed}[backgroundcolor=black!10,rightline=false,leftline=false,topline=false,bottomline=false,innerbottommargin=12pt,afterbreak={\hfill}]
\section*{Box 2.  Hippocampal replay}
Pyramidal neurons in the hippocampus exhibit spatial receptive fields: their firing rate increases by as much as tenfold when the animal is in a particular location [106]. Taken together, such ‘\textbf{place cells}’ have been proposed to form a cognitive map of an environment from which an animal may be able to plan routes, find shortcuts, and make other inference. As the animal traverses a room or a habitat, the sequence of increased firing rates of one place cell after another can provide a read-out of the animal’s trajectory through the environment [29]. Following earlier predictions [107,108], a series of studies in the 1990s showed that pairs of place cells that were coactive during behaviour (i.e., encoding overlapping or adjacent locations on a maze) became coactive again when the animal was taken away from the maze and left to rest or sleep in one place [31–34]. This reactivation of place cell pairs was above both the level of chance and the level of their coactivation during rest before exploring the environment; that is to say that the hippocampal trace of previous behaviour was being replayed during rest, when the animal was not running or exploring and the hippocampus was otherwise unengaged with the task of navigation (Figure 1).

Further research has shown that such replay extends outside the hippocampus to cortical [35–37] and limbic [38–40] brain areas, which are involved in processing non‐spatial information, suggesting a brain-wide phenomenon in which many facets of an experience, including sensory and reward-related properties, can be reactivated together.

In humans, the non‐invasive, non‐surgical experimental methods usually required for recording neural activity offer lower spatial and temporal resolution, making replay detection more difficult. Nevertheless, classifiers trained on human neural activity during a task show hippocampal reactivation of task representations during subsequent rest, with a bias towards replaying items that are highly rewarded and subsequently better remembered [109,110]. Replay has also been shown to selectively strengthen weaker memories [111] and re-evaluate state-action values for reinforcement learning [13], and with tentative evidence of hippocampal-to-cortical transfer of task-relevant information [112].

Evidence for the causal role of replay in memory consolidation has come from studies in which sharp-wave ripples in the hippocampus are either disrupted or prolonged. Hippocampal replay relies on synchronous excitation of large neuronal populations, which occurs during sharp-wave ripples [31]: distinctive, transient bursts of high-frequency oscillatory activity [113] that promotes the firing of a subset of pyramidal cells, resulting in replay sequences [26,27]. Disrupting the ripples, which also disrupts coincident replay events, results in slower spatial learning on timescales of minutes [4,5,10], and days [41,42]. Disrupting the replay event but not the ripple itself (technically a much harder feat) has also been shown to slow down learning [43]. Extending the duration of ripples, conversely, appears to increase replay and improve spatial memory [5]. Finally, studies that invoke or bias replay of some experiences over others can selectively improve the memory of those items [6–8,44]. The definitive evidence for a link between replay and memory consolidation would be performance improvement following the induction of a replay event from scratch. Such a test, however, is technically challenging, and to our knowledge has not been achieved experimentally so far.

\end{mdframed}}

\begin{figure*}
\centering
\includegraphics[width=.8\paperwidth]{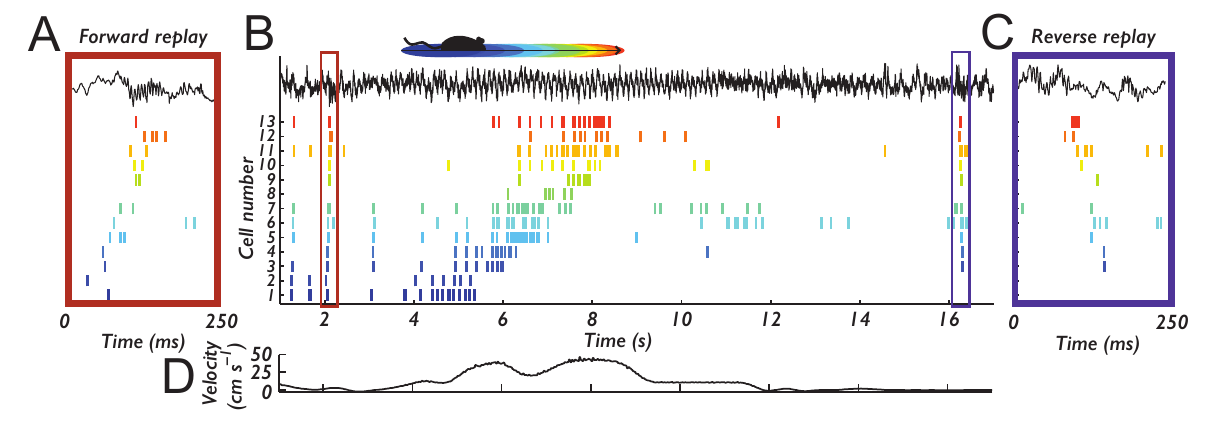}
\caption{\textbf{Replay of hippocampal place cells.} Replay of hippocampal place cells during a single lap of a linear track. (A) Spiking of place cells during a sharp-wave ripple, representing a forward replay. (B) Sequential spiking of place cells that encode successive locations on the track; colours represent the location on the track to which the cell is tuned. Black trace at the top shows concurrent local field potential. Red and blue boxes outline bursts of spiking, which are magnified in A and C, respectively. (C) Spiking of place cells during a sharp-wave ripple, representing a reverse replay. (D) Animal’s running velocity, including periods of immobility before and after the run. Figure adapted from [119].}
\label{fig:1}
\end{figure*}

A necessary element of the \textbf{deep Q network (DQN)} is that past trials are stored in a \textbf{memory buffer} and regularly played back to the network. Because the incoming training data depends on the agent’s previous actions, the
distribution of the training data is prone to shift as the agent’s \textbf{action policy} for how it behaves also shifts, leading to non‐stationary data distributions. Such temporal correlations between successive online learning trials can cause a phenomenon known as catastrophic forgetting, where weight parameters undergo changes that optimise for the most
recent gameplay at the cost of older gameplay; this results in behaviour learned from the previous task being rapidly overwritten by the agent’s new behaviour. Experience replay proved a crucial intervention to break temporal correlations between successive online learning trials, which stabilises learning and ultimately leads to much improved performance.

Experience replay itself was proposed in machine learning long before the deep reinforcement breakthrough, but it was only when experience replay and deep reinforcement learning were combined that closer parallels with replay in the brain emerged [14]. It can be argued that DQNs exemplify how artificial neural networks can be used as models of biological learning to test theories of how replay can support learning. Manipulating biological replay has largely been limited to broad disruption of replay patterns [4,5,10,41,42], typically during a relatively brief period immediately following learning, which is found to diminish learning. Whereas in DQNs the consequences of manipulations such as including, excluding, or tweaking replay have been examined more comprehensively [14,52], the effects of such manipulations in biological settings have been relatively less studied.

However, the differences between how experience replay is implemented algorithmically and how biological replay occurs physiologically merit further consideration. In the early experience replay method [14], the memory buffer consisted of past samples taken at random from the replay buffer. The memory buffer was set with an arbitrary capacity of the most recent one million trials; when full, the older samples in memory were replaced with new ones to maintain relatively recent training data. Several of these features of artificial experience replay (specifically the storage of exact copies of previous trials [53,54], the prioritisation of recent experience [55], uniform sampling from the memory buffer [56], and its fixed capacity [53]) have been developed further in recent years, and all of them can be said to be unrepresentative of biological replay to varying degrees. In the following sections, we highlight how advances in experience replay algorithms have come closer to replicating biological replay and the implications for replay as a cognitive mechanism.

\section*{Prioritised experience replay}
In principle, replaying activity corresponding to past trials effectively increases the training set, by supplementing
‘live’, online experiences with additional samples. Further, manipulating the set of replayed experiences (the replay buffer) can alter the statistics of this training set to promote more efficient learning. For example, given limited time or resources, prioritising more ‘important’ samples for replay can produce faster and more efficient learning of the encoded information. Depending on what is considered important for the animal or agent, this can be achieved by biasing replay towards rarer events to overcome the under-representation of such information online, towards information that is likely to be needed in the near future [4,11,37,38,57,58], or towards the most surprising or unexpected information [46,48].

\begin{figure*}
\centering
\includegraphics[width=.8\paperwidth]{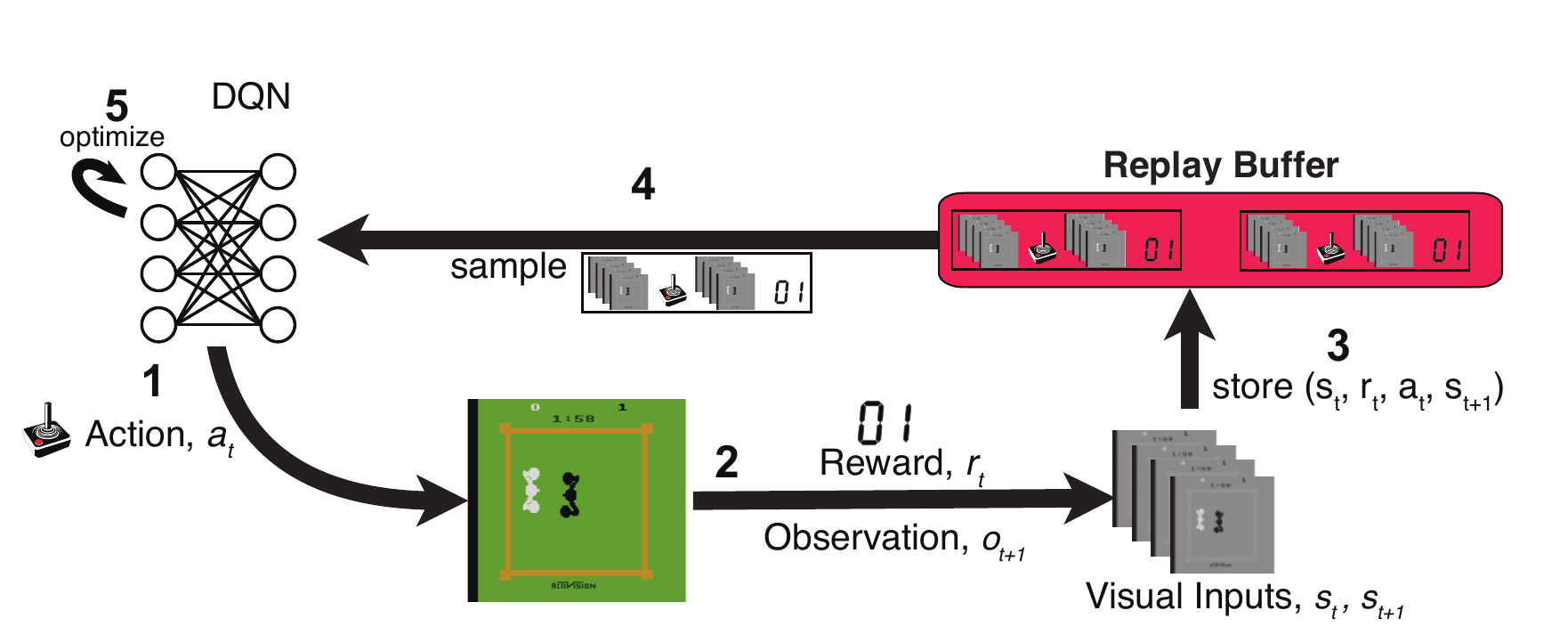}
\caption{\textbf{Deep Q network (DQN) with memory buffer.} Top: a DQN trained to play the Atari game Boxing. At every time-step $t$, the DQN outputs an action corresponding to a joystick movement (1), which causes the game to produce a new reward (game score) and a new observation (pixel values; 2). The observation is transformed into a series of four visual frames that make up the state, and the tuple of state, reward, action and subsequent state are stored in the replay buffer (3). These tuples are then sampled from the buffer and replayed to the DQN so that it optimises a function mapping state inputs to action outputs in order to maximise reward (4). Bottom: at each update, Q-values for a given pair of state $s$ and action $a$ from the sample are updated according to the difference between the observed value and expected value, where the observed value is the reward $r$ from the sample added to expected future reward $maxQ(s', a')$ discounted by a factor $\gamma$ and expected value is the previous Q-value of the state-action pair. Over repeated updates, the Q-values converge on an approximation of how a state-action pair (top node) maps onto possible subsequent states (white nodes) and actions (bottom nodes).}
\label{fig:1}
\end{figure*}

One approach is to select samples stored in the replay buffer by prioritising them according to the magnitude of the
\textbf{temporal-difference error} [56]: a measure of how unexpected a transition between temporally successive states has been. This prioritised replay has also been applied to computational models in neuroscience research, showing that considering rewards while accessing memory can lead to better planning and decision making [58]. Replaying surprising information is particularly beneficial, because examples where observed outcomes diverge from predictions indicate that the predictions need improving, so there is more to be gained from updating the weights that produced the erroneous prediction. Predictive coding, based on a mental model of the environment, is a feature of neural activity in the hippocampus and elsewhere [59] and offers an efficient way of learning by correcting for errors in predictions. Dopaminergic signals, which influence plasticity in the hippocampus and other brain areas that exhibit replay, have been argued to encode reward-prediction errors during reinforcement learning [60]. This means that there is a normative case to be made for the brain to prioritise neural activity associated with errors and there are plausible \textbf{neuromodulatory} mechanisms by which this might be instantiated. Relatedly, it has been shown that using DQNs with replay biased by prediction errors can speed up learning [56].

In the biological context, prioritised experience is exemplified by the finding in rats that there is greater replay immediately following reward in a reinforcement learning task than following the absence of reward [46] or following reward that does not depend on reinforcement learning [61], suggesting a bias towards (positive) reward-prediction error. Biological prioritised experience replay is believed to rely on neuromodulatory influences in networks of neurons [62]. The association between reward and replay had been observed not only immediately after a trial during wakefulness but also during subsequent sleep; however, these findings are consistent with several possible computational roles for reward in replay [63]. More generally, replay has been found to be biased by aversive as well as rewarding outcomes [40,64], suggesting that the brain does engage in methods of prioritising some replay events over others. In addition to carefully designed experiments to tease out how replay contributes to reinforcement learning in animals [63], directly examining the effects of prioritised replay on a deep neural network can be an important proof of principle for how different replay regimes affect learning.

Overall, this is a research area where AI, we would argue, could be a good model for biological learning, because the prediction-error theory of the midbrain dopaminergic circuit has been successful historically in explaining a wide range of experimental findings, and several parts of that circuit have been implicated in replay [13,36–39]: hippocampus and prefrontal cortex for encoding states [65,66], striatum for action values [67,68], and VTA for reward-prediction errors [69,70]. Therefore, it provides a suitable framework for integrating hypotheses about replay, reinforcement learning, dopaminergic signalling, and temporal-difference errors.

Other reviews have covered in greater depth the suitability of deep neural networks as models of neuronal function [71–74]. There are fundamental differences between biological and artificial networks, but, in various domains, DNNs can be useful as abstract neural models especially for manipulations that are not easy to test biologically; for example, to model how different neural architectures and cost functions can support learning and performance. This is exemplified by convolutional neural networks and long short-term memory architectures, which have been successfully used as models for visual processing and short-term memory, respectively [75]. For replay, manipulating the rules for prioritising replay in deep neural networks can offer insight into how it supports cognitive functions. Where deep neural networks diverge from biological networks, there is an opportunity to identify the necessary elements for recreating desired features. In-depth investigation of how such differences affect information processing in both biological and artificial neural networks has the potential to reveal the key computational principles underlying learning in these systems.

For example, it may be important to establish some quantification of phenomena such as interleaving of online and offline episodes. Quantifying replay over the course of learning in animal studies is difficult for several reasons, among which is a lack of consistency in how replay is defined, measured, and reported. However, the majority of decodable replay events immediately following an episode or trial are repetitions of the same hippocampal activity rather than replays of a different episode [45], a proportion that decreases with learning [76]. Specifically, 0.4–2 \textbf{hippocampal replays} of a familiar, low-reward and low-prediction-error episode per online trial have been reported [77–79], rising to 1–4 replays per online trial for higher-reward episodes [45,76,77,79] and as many as 9 replays per online trial for episodes with high reward-prediction error [77]. This compares to, for example, a mini-batch of 32 samples from the replay buffer for each online trial in a DQN [14]. Similarly, before initiating an action, the majority of decodable replay events reflect the upcoming trajectory [12,45] (but see [79]), a proportion that increases with learning [76].

Does this mean replays during learning are mostly massed repetitions and not interleaved? This remains to be seen. Some place cells are selective not just for spatial location but also temporal or behavioural context [45,77,78,80]; whether they differentially participate in replay events to interleave different contexts for the same spatial trajectory is not clear. In addition, what is replayed in prefrontal cortex appears to be mostly independent of what is replayed in hippocampus [81], which suggests that interleaving may happen at a higher cortical, rather than hippocampal, level. Finally, decodable replay events typically coincide with half or less of sharp-wave ripple events, which means the remaining ripples may contain undetected replays of hippocampal activity encoding other environments or tasks [47] (but see [79]).

Biological replay continues for hours after training at a slowly decaying rate [82], so sleep or extended rest may offer an opportunity for increasingly interleaved replay of these new episodes with older ones [47]. However, what is replayed, apart from the novel experience, is largely unknown, because studies typically record a single task in one environment to which replayed activity can be decoded. Ambitious experiments with several environments or reinforcement learning tasks, varying in their similarity to each other and recorded over long timescales, would be beneficial for testing predictions about how replay is interleaved.

\section*{Memory buffer as a limitation of replay}
Uniform and prioritised experience replay have proved useful for learning individual, relatively deterministic tasks such as video games. But as the AI field becomes more ambitious in its application domains, the tasks to which models are applied become more complex and less constrained, necessitating more sophisticated solutions. One example is hindsight experience replay [83], in which the learning agent breaks down complicated goals into simpler ones by redefining its goal (target state) retrospectively when replaying past experience. This flexibility makes it easier to learn from sparse rewards and allows learning of many policies in parallel, supporting efficient learning for robotics where there are many more degrees of freedom in how an agent can interact with the physical world than a joystick-controlled video game with limited screen resolution. Even this has limitations when the tasks to be learned are too distinct from each other.

\afterpage{\begin{mdframed}[backgroundcolor=black!10,rightline=false,leftline=false,topline=false,bottomline=false,innerbottommargin=12pt,afterbreak={\hfill}]
\section*{Box 3.  Proposed computational functions of replay}

Replay has been proposed to serve a variety of cognitive and network functions for learning, some of which depend biologically on brain area and sleep–wake state (during behaviour, extended rest, rapid eye movement (REM) sleep, or non-REM sleep). The theories and perspectives outlined briefly in the following bullet points are not mutually exclusive, and some of them are not unique to reinforcement learning, but they suggest how replay can support learning from individual rewarded episodes.

\begin{itemize}
  \item Consolidation of new memories rapidly encoded by a fast learner (the hippocampus) into long-term storage in a slow learner (the cortex) [57]. Memory encoding can form rapidly in the hippocampus but takes longer to be integrated into cortical representations (but see [114,115]). Replay serves as additional training to supplement online learning, in order to strengthen cortical representations [116].
  
  \item Generalising across episodes. Representations formed quickly and sparsely, for example in the hippocampus, enable pattern separation, which is beneficial for one-shot learning or retention of individual episodes of experience, but poorly suited to generalising across episodes. By contrast, a slower-learning cortex can integrate multiple episodes and encode statistical regularities between them, but takes many examples to achieve this (but see[117]). Replay serves as additional training for the cortex from individual episodes initiated by the hippocampus.
  
  \item Preventing catastrophic interference. Gradual interleaving of online and offline information regularises synaptic changes or weight changes to ensure that network representations of older information are not supplanted by those of newer information [30,57].
  
  \item Stabilising learning. Experiences close in time tend to be correlated, which can result in large, fluctuating weight changes. Interleaving uncorrelated samples constrains weight changes for more stable learning [57,118].
  
 \end{itemize}

\end{mdframed}}

A further limitation is that memory buffers in deep reinforcement learning are often designed to contain one million transitions, an arbitrary hyperparameter that is commonly untuned. With more complex tasks, a limited memory buffer cannot contain a representative set of samples from a very large state-space: some will fail to be stored in the memory buffer, so the network is at risk of converging on solutions that are not appropriate for these forgotten experiences. This storage issue is particularly problematic for continual learning when many tasks must be learned in sequence, which is becoming an important problem as AI progresses towards ever more complex challenges.

One response to this challenge is to innovate ways to prioritise the samples that are stored in the memory buffer, contrasting with the simpler approach of just making the buffer bigger. Techniques have been developed to penalise, limit, or freeze weight changes in the network that were important for one task when learning a new additional task. For example, with elastic weight consolidation, the learnt connections that are important for one task but not another can be restricted to limit weight changes, ensuring that new learning is encoded by a different set of connections in the same network [84]. To some degree this resembles the largely non‐overlapping cortical neuronal representations that are enforced when successive tasks are learnt in quick succession, which has been shown to prevent interference [85], although this is not characteristic of hippocampus [86], and over longer timescales and larger spatial scales ‘representational drift’ achieves a similar effect by different mechanisms [80]. Variations of this approach mean that the replay buffer can either be abolished altogether, or used only to store a limited selection of recent samples (such as from the current task) [87–91].

Although these approaches prevent catastrophic forgetting, often without needing a replay buffer, they often assume a series of discrete, well-defined tasks with parameters that do not need to be updated once learned [92]. This may not be suitable for autonomous agents that face continuous, unsupervised learning in the real world. In addition, reducing the overlap between representations of different tasks in the network disregards the potential benefits of transfer learning, in which shared representations can speed up learning on new, similar tasks.

How can replay be used to support continual learning? To some extent, maintaining the memory buffer as a representative sample of tasks past and present can mitigate catastrophic forgetting [55], but eventually the trade-off remains between a large computationally costly memory buffer and remembering insufficient samples. Another ethical consideration is that explicitly storing training data may also violate privacy for some applications. Given that the mammalian brain achieves both continual learning and transfer learning with ease despite limits on resources, again neuroscience has major potential to offer inspiration for solving this problem: converging new evidence suggests that the hippocampus is capable of generating its own samples to train from.

\section*{Generative replay}

\begin{figure*}
\centering
\includegraphics[width=.6\paperwidth]{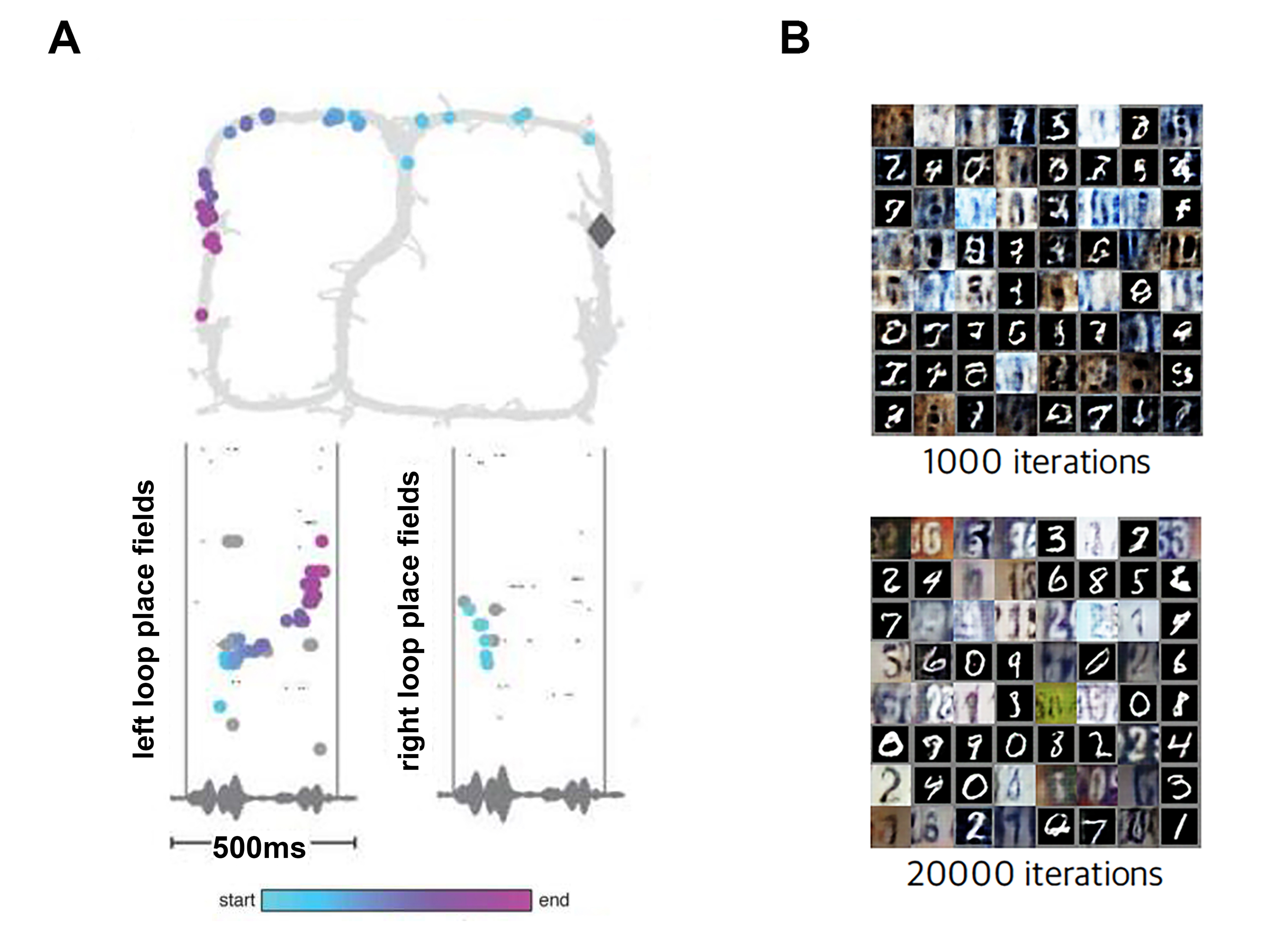}
\caption{\textbf{Biological and artificial generative replay.} (A) Hippocampal replay sequences have been found to reflect paths through the environment never taken by the animal, as demonstrated by decoding of locations during a replay event along the top of the maze when the animal had only explored it in a figure-of-eight pattern. Bottom: spiking of hippocampal cells that have place fields on the left and right loops of the maze, respectively, during a replay event. Top: decoded position during replay event. Colour indicates temporal order in the replay sequence. (B) Images generated by a generative adversarial network trained to reproduce handwritten MNIST digits and Street View House Numbers, instead of explicitly storing copies of such images in a memory buffer as in standard experience replay. Sample examples from early (top) and later in training (bottom) are shown; as training progresses, the samples produced by the generator increasingly resemble real images and are increasingly recognisable as numbers. Panels (A) and (B) adapted from [53,94], respectively.}
\label{fig:1}
\end{figure*}

Brains are not structured in a way that separates the processing network away from the learning network that stores exact copies of past trials. Moreover, neuronal firing in brains is inherently stochastic. The result is that no two patterns of activity, either the original online activity encoding the novel experience, or any subsequent replay, are exactly the same. In the context of spatial navigation, sequences of activity during replay events can reflect random walks through the cognitive map [93], encoding paths through the environment that have never been observed directly by an animal, including reverse trajectories in a one-way system, shortcuts, and blocked-off paths that were seen but not taken [94,95] (Figure 3). One can interpret this as if the hippocampus generates its own samples (its own training data) from a minimal stored model.

Deep neural networks can generate their own samples too, although so far this has been achieved primarily in supervised classification tasks with relatively simple datasets and with minimal applications for reinforcement learning. Typically, a generative adversarial network, autoencoder, or restricted Boltzmann machine is trained to generate data with the same parameter distributions as incoming data. The generated data is then replayed to the (main) solver model, so that even as new tasks are learned, the cumulative input distribution of all previous tasks continue to be replayed with minimal storage burden.

This approach has been shown to work well for classifying the MNIST database of handwritten digits learnt in sequence without succumbing to catastrophic forgetting [53] (Figure 3). So far, implementations of generative replay usually train a separate network for each task, assuming well-isolated tasks. The challenge for efficient deep generative replay is training a generator that is flexible enough to be scalable to many tasks and complex inputs.

Future research on this topic could, in principle, apply generative replay to a broader range of reinforcement learning tasks and extend it as a model of biological replay. Hippocampal replay is influenced by neuromodulation both at the time of encoding and during replay [96], which could serve to bias the generative mechanism towards goal states or high-error transitions, or otherwise alter the parameters of the generated samples in a way that enhances reinforcement learning. There is also evidence of a bidirectional process whereby cortical activity can prompt or select replay patterns in hippocampus [8,44,97], in addition to the hippocampus replaying information to the cortex and other brain areas. These are under‐explored in modelling hippocampal replay and may also serve to improve the performance of deep generative replay for more complex tasks. There are many ways in which replayed information need not reflect online updates: low-dimensional representations, top-down influences, biasing towards salient information, and different update or plasticity rules could all be used to improve replay algorithms. For example, work on deep generative replay suggests that freezing weights in initial layers of the network and replaying low-dimensional information may reduce the computational burden [54]. An improved understanding of how replay patterns propagate through cortical and subcortical networks may inspire further refinements to this approach.

Finally, the additional flexibility that comes with generative replay may offer other benefits: one example of non‐stereotyped replay is that hidden structures can be inferred from information presented in a disjointed way. This pertains not only to finding shortcuts in spatial environments [94], but, more abstractly, to replaying sequences of images in the appropriate order [9]. Inferring hidden structures from segmented experience may not be possible with an explicit memory buffer, but such forms of biologically inspired replay might allow deep neural networks to gain more insight from their training than current methods can support. A possible role for replay in transfer learning, in which knowledge of past tasks is leveraged to improve learning on new tasks, is yet to be investigated in either neuroscience or machine learning, but may offer both improved algorithms and better insight into cognitive processes.

\section*{Future directions}
Recent experimental innovations allow recording of broader areas of the brain over longer timescales to capture more neural data [9,80,97]. This richer data offers possibilities for exploring how replay spreads throughout the brain [8,81] and over time [47,53,82], but will need strong theory to guide the boundless possibilities for analysis. Meanwhile, reinforcement learning in AI is increasingly taking advantage of more complex and brain-inspired representations of the world to meet the challenges of complex, noisy, and non‐stationary applications, including hierarchies of learners [98], and distributions of values [99]. Increasingly sophisticated task designs in experimental neuroscience may advance current understanding of replay of inferred structure [9], generalisations [81], interleaved episodes [47], and other ways of organising information to support complex cognition. In turn, these biological insights could help improve the use of past samples in machine learning to support transfer, continual and hierarchical reinforcement learning. While this recent focus on increasing the quantity of data is valuable, understanding the role of replay in complex cognitive processes will require more sophisticated and carefully designed behavioural tasks than are now available.

\afterpage{\begin{mdframed}[backgroundcolor=black!10,rightline=false,leftline=false,topline=false,bottomline=false,innerbottommargin=12pt,afterbreak={\hfill}]
\section*{Outstanding questions}

To what extent does replay reflect individual episodes versus aggregated statistics in the brain and how does the replay of individual versus generalised information support learning and consolidation in both biology and AI?

Replay in deep reinforcement learning often starts at the input layer and propagates to higher layers; in the mammalian brain it is unclear how different brain areas interact during replay events. Can replay originate in lower layers (corresponding to sensory cortices, in brains), higher layers (hippocampus, prefrontal cortex), and output layers (striatum)? How does it propagate through the network and what influence does this have on learning?

How does replay of newer versus older information, and overlapping versus distinct representations, support learning and consolidation?

How can replay of similar information benefit transfer learning?

Are the update rules different for online experience versus replay? Are the update rules different for replay during wake versus sleep and replay during rapid eye movement (REM) versus non-REM sleep?

How can generative replay be used to support continual learning of complex tasks?

\end{mdframed}}

Other developments include encoding states of the world as successor representations [100], a computational structure that represents states in terms of where they might lead to (i.e., expected occupancies of future states). Successor representations make learning more efficient, generalise better in multi‐task settings [101], and can be used as a model of hippocampal place cells [59]. Replaying the successor representations themselves, rather than pixel observations as is typical in experience replay [14], could further support learning in complex settings, not only improving machine learning algorithms but also shedding light on the function of hippocampal replay.

Beyond gaining insights into optimisation of cognitive processes, closer integration of research on replay for reinforcement learning in AI and neuroscience could lead to new avenues for understanding disease aetiology and symptoms. Reinforcement learning is altered, for instance, in patients with Parkinson’s disease and schizophrenia (conditions that are characterised by altered dopamine levels), both during learning itself and in the consolidation period of hours or days afterwards [102,103]. Future models of replay for reinforcement learning that explicitly relate to the various neural circuits affected by such conditions may help to explain how they give rise to the associated cognitive changes.

\section*{Concluding remarks}
Deep neural networks and brains are examples of artificial and natural intelligent systems that solve computationally related problems involving reinforcement learning in complex environments. Despite significant fundamental differences between biological and artificial networks, replay appears to reflect convergence on comparable mechanisms: both biologically and artificially intelligent systems harness replay of past experience to stabilise, consolidate, generalise from, and accelerate learning. In neuroscience, questions remain about how replay is initiated and propagated through cortical and subcortical networks and what influence this has on learning. By addressing these problems in simulated and real-world settings, computational models can play important roles in making testable predictions of how the brain solves these problems (see Outstanding questions). Deep neural networks are an example of such a computational model that underlie complex applications of AI in computer vision and other domains while also being biologically plausible in some aspects, making them invaluable for manipulating aspects of replay that are difficult to control experimentally to reveal how it supports learning and memory processes. In particular, experimental studies are restricted to learning and memory processes that take place over timescales of seconds to weeks; therefore, hypotheses about lifelong learning may be more easily developed and tested in deep neural networks. Conversely, as research on replay throughout the brain develops, it may hold keys to solving some of the biggest challenges in AI today, including continual learning where algorithms fall short of what brains can do with ease [104]. A more ambitious long-term goal for AI, with prospects of its feasibility being debated, is the design of an artificial general intelligence that can tackle virtually any task in the grasp of humans.

\afterpage{\begin{mdframed}[backgroundcolor=black!10,rightline=false,leftline=false,topline=false,bottomline=false,innerbottommargin=12pt,afterbreak={\hfill}]
\section*{Glossary}

\textbf{Action policy:} behaviour learned by a reinforcement learning agent that determines the actions to be taken given
observations about the current state.

\textbf{Backpropagation:} learning algorithm for updating the weights in a deep neural network, in which the gradients of the error function are calculated for each layer sequentially, starting with the last layer.

\textbf{Catastrophic interference:} (or forgetting); when new learning in a network causes dramatic changes that result in the loss of
previously acquired, stable associations.

\textbf{Deep Q network (DQN):} deep neural network that performs Q-learning.

\textbf{Deep reinforcement learning:} reinforcement learning algorithms implemented in deep neural networks, characterised by an input later, an output layer, and at least one intermediate hidden layer.

\textbf{Dopamine:} neuromodulator released by (among other brain regions) the basal forebrain. Dopamine has been proposed to function as a reward-prediction error signal throughout the brain.

\textbf{Experience replay:} technique of sampling from past experiences stored in a memory buffer, and replaying them to the network.

\textbf{Experience tuple:} the data from a single episode that is stored in a memory buffer to be replayed later; typically, the state of the environment, the action taken by the agent, the resulting reward, and the subsequent state of the 
environment.

\textbf{Hippocampal replay:} reinstatement of neural activity that encodes a previous experience in the hippocampus during rest
and sleep.

\textbf{Memory buffer:} storage of experience tuples during the learning of a task, which can later be sampled from and replayed.

\textbf{Memory consolidation:} stabilisation of a recent memory into the neural circuit so that it is retained long-term, through chemical and structural plasticity processes.

\textbf{Neuromodulation:} chemical transmission between neurons that diffuses over a broad area, to regulate the activity of a large number of neurons.

\textbf{Non-REM sleep:} light and deep sleep stages, excluding rapid‐ eye‐ movement (REM) sleep, characterised by synchronous patterns of neural activity. Most replay has been observed in non-REM sleep, and it is unclear to what extent the quantity, attributes, and computational roles of replay differ between REM and non-REM sleep.

\textbf{Place cell:} neuron found in the hippocampus that fires preferentially when the animal is in a particular place.

\textbf{Plasticity:} changes in neural circuits, particularly the strengthening and weakening of synaptic connections.

\textbf{Q-learning:} a model-free reinforcement learning algorithm and extension of temporal-difference learning, for optimising the policy of selecting actions in any given state.

\textbf{Reinforcement learning:} learning to act to maximise expected rewards, an area of study in both psychology and machine
learning.

\textbf{State transitions:} the possibility or probability of one state leading to another state in a single step; in spatial navigation tasks this depends on the topology of the environment, including physical distance between states and barriers between them.

\textbf{Temporal-difference error:} reinforcement learning algorithm that learns the values of states by minimising the difference in predicted value between temporally successive states.

\end{mdframed}}

\subsection*{Acknowledgements}

This research was funded in part by a Wellcome Senior Research Fellowship in Basic Biomedical Science (grant number 202810/Z/16/Z to M.W.J.), a Wellcome Neural Dynamics PhD studentship (grant number 109070/Z/15/A to E.R.), and a Leverhulme Trust Research Leadership Award (grant number RL-2016-39 to N.L.). R.C. was supported by Unifying Neuroscience and Artificial Intelligence - Québec (UNIQUE), Fonds de recherche du Québec (FRQNT) and Natural Sciences and Engineering Research Council of Canada (NSERC). For the purpose of Open Access, the authors have applied a CC BY public copyright licence to any Author Accepted Manuscript version arising from this submission.

\subsection*{Declaration of interests}
The authors declare no competing interests in relation to this work.

\section*{References}

\begin{enumerate}

    \item R. S. Sutton and A. G. Barto, Introduction to Reinforcement Learning. Cambridge: The MIT Press, 1998.

    \item L.-J. Lin, “Self-improving reactive agents based on reinforcement learning, planning and teaching,” Mach. Learn., vol. 8, no. 3–4, pp. 293–321, May 1992.
    
    \item H. F. Ólafsdóttir, D. Bush, and C. Barry, “The Role of Hippocampal Replay in Memory and Planning.,” Curr. Biol., vol. 28, no. 1, pp. R37–R50, 2018.
    
    \item F. Michon, J.-J. Sun, C. Y. Kim, D. Ciliberti, and F. Kloosterman, “Post-learning Hippocampal Replay Selectively Reinforces Spatial Memory for Highly Rewarded Locations,” Curr. Biol., vol. 29, no. 9, pp. 1436–1444, 2019.
    
    \item A. Fernández-Ruiz, A. Oliva, E. Fermino de Oliveira, F. Rocha-Almeida, D. Tingley, and G. Buzsáki, “Long-duration hippocampal sharp wave ripples improve memory.,” Science, vol. 364, no. 6445, pp. 1082–1086, 2019.
    
    \item B. Rasch, C. Büchel, S. Gais, and J. Born, “Odor cues during slow-wave sleep prompt declarative memory consolidation,” Science, vol. 315, no. 5817, pp- 1426-1429, 2007.
    
    \item D. C. Barnes and D. A. Wilson, “Slow-Wave Sleep-Imposed Replay Modulates Both Strength and Precision of Memory,” Science, vol. 34, no. 15, pp.5134-5142, 2014.
    
    \item G. Rothschild, E. Eban, and L. M. Frank, “A cortical-hippocampal-cortical loop of information processing during memory consolidation.,” Nat. Neurosci., vol. 20, no. 2, pp. 251–259, 2017.
    
    \item Y. Liu, R. J. Dolan, Z. Kurth-Nelson, and T. E. J. Behrens, “Human Replay Spontaneously Reorganizes Experience,” Cell, vol. 178, no. 3, p. 640–652.e14, 2019.
    
    \item S. P. Jadhav, C. Kemere, P. W. German, and L. M. Frank, “Awake Hippocampal Sharp-Wave Ripples Support Spatial Memory,” Science, vol. 336, no. 6087, pp. 1454–1458, 2012.
    
    \item A. A. Carey, Y. Tanaka, and M. A. A. van der Meer, “Reward revaluation biases hippocampal replay content away from the preferred outcome,” Nat. Neurosci., vol. 22, no. 9, pp. 1450–1459, 2019.
    
    \item H. F. Ólafsdóttir, F. Carpenter, and C. Barry, “Task Demands Predict a Dynamic Switch in the Content of Awake Hippocampal Replay,” Neuron, vol. 96, no. 4, p. 925–935.e6, 2017.
    
    \item I. Momennejad, A. R. Otto, N. D. Daw, and K. A. Norman, “Offline replay supports planning in human reinforcement learning,” Elife, vol. 7, 2018.
    
    \item V. Mnih et al., “Human-level control through deep reinforcement learning,” Nature, vol. 518, no. 7540, pp. 529–533, 2015.
    
    \item D. Silver et al., “Mastering the game of Go with deep neural networks and tree search,” Nature, vol. 529, no. 7587, pp. 484–489, 2016.
    
    \item H. R. Joo and L. M. Frank, “The hippocampal sharp wave–ripple in memory retrieval for immediate use and consolidation,” Nat. Rev. Neurosci., vol. 19, no. 12, pp. 744–757, 2018.
    
    \item D. J. Foster, “Replay Comes of Age,” Annu. Rev. Neurosci., vol. 40, pp. 581–602, 2017.
    
    \item G. Rothschild, “The transformation of multi-sensory experiences into memories during sleep,” Neurobiol. Learn. Mem., vol. 160, pp. 58–66, 2019.
    
    \item B. E. Pfeiffer, “The content of hippocampal ‘replay,’” Hippocampus, vol. 30, no. 1, pp. 6–18, 2020.
    
    \item R. Stickgold, “Sleep-dependent memory consolidation,” Nature, vol. 437, no. 7063, pp. 1272–1278, 2005.
    
    \item J. H. L. P. Sadowski, M. W. Jones, and J. R. Mellor, “Sharp-Wave Ripples Orchestrate the Induction of Synaptic Plasticity during Reactivation of Place Cell Firing Patterns in the Hippocampus,” Cell Rep., vol. 14, no. 8, pp. 1916–1929, 2016.
    
    \item C. J. Behrens, L. P. van den Boom, L. de Hoz, A. Friedman, and U. Heinemann, “Induction of sharp wave–ripple complexes in vitro and reorganization of hippocampal networks,” Nat. Neurosci., vol. 8, no. 11, pp. 1560–1567, 2005.
    
    \item H. Norimoto et al., “Hippocampal ripples down-regulate synapses.,” Science, vol. 359, no. 6383, pp. 1524–1527, 2018.
    
    \item E. V. Lubenov and A. G. Siapas, “Decoupling through Synchrony in Neuronal Circuits with Propagation Delays,” Neuron, vol. 58, no. 1, pp. 118–131, 2008.
    
    \item L. L. Colgin, D. Kubota, Y. Jia, C. S. Rex, and G. Lynch, “Long-term potentiation is impaired in rat hippocampal slices that produce spontaneous sharp waves,” J. Physiol., vol. 558, no. 3, pp. 953–961, 2004.
    
    \item Z. Nádasdy, H. Hirase, and A. Czurkó, “Replay and time compression of recurring spike sequences in the hippocampus,” J. Neurosci.,vol. 19, no. 21, pp. 9497-507, 1999.
    
    \item A. K. Lee and M. A. Wilson, “Memory of Sequential Experience in the Hippocampus during Slow Wave Sleep,” Neuron, vol. 36, no. 6, pp. 1183–1194, 2002.
    
    \item M. Yoshida, B. Knauer, and A. Jochems, “Cholinergic modulation of the CAN current may adjust neural dynamics for active memory maintenance, spatial navigation and time-compressed replay,” Front. Neural Circuits, vol. 6, no. 10, 2012.
    
    \item K. Diba and G. Buzsáki, “Forward and reverse hippocampal place-cell sequences during ripples,” Nat. Neurosci., vol. 10, no. 10, pp. 1241–1242, 2007.
    
    \item J. L. McClelland, B. L. McNaughton, and R. C. O’Reilly, “Why there are complementary learning systems in the hippocampus and neocortex: Insights from the successes and failures of connectionist models of learning and memory.,” Psychol. Rev., vol. 102, no. 3, pp. 419–457, 1995.
    
    \item M. McCloskey and N. J. Cohen, “Catastrophic Interference in Connectionist Networks: The Sequential Learning Problem,” Psychol. Learn. Motiv. - Adv. Res. Theory, vol. 24, no. C, pp. 109–165, 1989.
    
    \item M. Karlsson and L. Frank, “Awake replay of remote experiences in the hippocampus,” Nat. Neurosci., vol. 10, pp. 100-107, 2007.
    
    \item D. Ji and M. Wilson, “Coordinated memory replay in the visual cortex and hippocampus during sleep,” Nat. Neurosci., vol. 10, pp. 100-107, 2007.
    
    \item D. R. Euston, M. Tatsuno, and B. L. McNaughton, “Fast-Forward Playback of Recent Memory Sequences in Prefrontal Cortex During Sleep,” Science, vol. 318, no. 5853, pp. 1147-1150, 2007.
    
    \item A. Peyrache, M. Khamassi, and K. Benchenane, “Replay of rule-learning related neural patterns in the prefrontal cortex during sleep,” Nature, vol. 12, pp. 919-926, 2009.

    \item M. A. Wilson and B. L. McNaughton, “Dynamics of the hippocampal ensemble code for space,” Science, vol. 261, no. 5124, pp. 1055–1058, 1993.

    \item M. A. Wilson and B. L. McNaughton, “Reactivation of hippocampal ensemble memories during sleep.,” Science, vol. 265, no. 5172, pp. 676–9, 1994.

    \item W. E. Skaggs and B. L. McNaughton, “Replay of neuronal firing sequences in rat hippocampus during sleep following spatial experience.,” Science, vol. 271, no. 5257, pp. 1870–3, 1996.

    \item H. S. Kudrimoti, C. A. Barnes, and B. L. McNaughton, “Reactivation of hippocampal cell assemblies: effects of behavioral state, experience, and EEG dynamics.,” J. Neurosci., vol. 19, no. 10, pp. 4090–101, 1999.

    \item Y.-L. Qin, B. L. Mcnaughton, W. E. Skaggs, and C. A. Barnes, “Memory reprocessing in corticocortical and hippocampocortical neuronal ensembles,” Philos. Trans. R. Soc. B Biol. Sci., vol. 352, no. 1360, pp. 1525–1533, 1997.

    \item C. M. A. Pennartz, E. Lee, J. Verheul, P. Lipa, C. A. Barnes, and B. L. McNaughton, “The Ventral Striatum in Off-Line Processing: Ensemble Reactivation during Sleep and Modulation by Hippocampal Ripples,” J. Neurosci., vol. 24, no. 29, pp. 6446–6456, 2004.

    \item S. N. Gomperts et al., “VTA neurons coordinate with the hippocampal reactivation of spatial experience,” Elife, vol. 4, pp. 321–352, 2015.

    \item G. Girardeau, I. Inema, and G. Buzsáki, “Reactivations of emotional memory in the hippocampus–amygdala system during sleep,” Nat. Neurosci., vol. 20, no. 11, pp. 1634–1642, 2017.

    \item Z. Nádasdy, H. Hirase, A. Czurkó, J. Csicsvari, and G. Buzsáki, “Replay and time compression of recurring spike sequences in the hippocampus.,” J. Neurosci., vol. 19, no. 21, pp. 9497–507, 1999.

    \item G. Girardeau, K. Benchenane, S. I. Wiener, G. Buzsáki, and M. B. Zugaro, “Selective suppression of hippocampal ripples impairs spatial memory,” Nat. Neurosci., vol. 12, no. 10, pp. 1222–1223, 2009.

    \item V. Ego-Stengel and M. A. Wilson, “Disruption of ripple-associated hippocampal activity during rest impairs spatial learning in the rat.,” Hippocampus, vol. 20, no. 1, pp. 1–10, 2010.

    \item I. Gridchyn, P. Schoenenberger, J. O’Neill, and J. Csicsvari, “Assembly-Specific Disruption of Hippocampal Replay Leads to Selective Memory Deficit,” Neuron, vol. 106, no. 2, p. 291–300.e6, 2020.

    \item D. Bendor and M. A. Wilson, “Biasing the content of hippocampal replay during sleep.,” Nat. Neurosci., vol. 15, no. 10, pp. 1439–44, 2012.

    \item H. Xu, P. Baracskay, J. O’Neill, and J. Csicsvari, “Assembly Responses of Hippocampal CA1 Place Cells Predict Learned Behavior in Goal-Directed Spatial Tasks on the Radial Eight-Arm Maze,” Neuron, vol. 101, no. 1, p. 119–132.e4, 2019.

    \item A. C. Singer and L. M. Frank, “Rewarded Outcomes Enhance Reactivation of Experience in the Hippocampus,” Neuron, vol. 64, no. 6, pp. 910–921, 2009.

    \item C. S. Lansink, P. M. Goltstein, J. V. Lankelma, R. N. J. M. A. Joosten, B. L. McNaughton, and C. M. A. Pennartz, “Preferential Reactivation of Motivationally Relevant Information in the Ventral Striatum,” J. Neurosci., vol. 28, no. 25, 2008.

    \item S. J. Gershman, A. B. Markman, and A. R. Otto, “Retrospective Revaluation in Sequential Decision Making: A Tale of Two Systems,” J. Exp. Psych. Gen., vol. 143, no. 1, p. 182, 2012.

    \item D. Hassabis, D. Kumaran, C. Summerfield, and M. Botvinick, “Neuroscience-Inspired Artificial Intelligence,” Neuron, vol. 95, no. 2, pp. 245–258, 2017.

    \item W. Fedus et al., “Revisiting Fundamentals of Experience Replay,” Proceedings of the 37th International Conference on Machine Learning, vol. 119, pp. 3061-3071, 2020.

    \item H. Shin, J. K. Lee, J. Kim, and J. Kim, “Continual Learning with Deep Generative Replay,” arXiv preprint arXiv:1705.08690, 2017.

    \item G. M. van de Ven, H. T. Siegelmann, and A. S. Tolias, “Brain-inspired replay for continual learning with artificial neural networks,” Nat. Commun., vol. 11, no. 1, 2020.

    \item D. Isele and A. Cosgun, “Selective Experience Replay for Lifelong Learning,” Proceedings of the AAAI Conference on Artificial Intelligence, vol. 32, no. 1, 2018.

    \item T. Schaul, J. Quan, I. Antonoglou, D. Silver, and G. Deepmind, “Prioritized Experience Replay,” arXiv preprint arXiv:1511.05952, 2015.

    \item M. G. Mattar and N. D. Daw, “Prioritized memory access explains planning and hippocampal replay,” Nat. Neurosci., vol. 21, no. 11, pp. 1609-1617, 2018.

    \item K. L. Stachenfeld, M. M. Botvinick, and S. J. Gershman, “The hippocampus as a predictive map,” Nat. Neurosci., vol. 20, no. 11, pp. 1643–1653, 2017.

    \item M. Watabe-Uchida, N. Eshel, and N. Uchida, “Neural Circuitry of Reward Prediction Error,” Annu. Rev. Neurosci., vol. 40, pp. 373–394, 2017.

    \item C.-T. Wu, D. Haggerty, C. Kemere, and D. Ji, “Hippocampal awake replay in fear memory retrieval,” Nat. Neurosci., vol. 20, no. 4, pp. 571–580, 2017.

    \item R. C. Wilson, Y. K. Takahashi, G. Schoenbaum, and Y. Niv, “Orbitofrontal cortex as a cognitive map of task space,” Neuron, vol. 81, no. 2, pp. 267–279, 2014.

    \item A. M. Wikenheiser and G. Schoenbaum, “Over the river, through the woods: Cognitive maps in the hippocampus and orbitofrontal cortex,” Nature Reviews Neuroscience, vol. 17, no. 8, pp. 513–523, 2016.

    \item K. Samejima, Y. Ueda, K. Doya, and M. Kimura, “Representation of action-specific reward values in the striatum,” Science, vol. 310, no. 5752, pp. 1337–1340, 2005.

    \item B. Lau and P. W. Glimcher, “Value Representations in the Primate Striatum during Matching Behavior,” Neuron, vol. 58, no. 3, pp. 451–463, 2008.

    \item W. Schultz, “Predictive Reward Signal of Dopamine Neurons,” J. Neurophysiol., vol. 80, no. 1, pp. 1–27, 1998.

    \item J. P. O’Doherty, P. Dayan, K. Friston, H. Critchley, and R. J. Dolan, “Temporal Difference Models and Reward-Related Learning in the Human Brain,” Neuron, vol. 38, no. 2, pp. 329–337, 2003.

    \item A. H. Marblestone, G. Wayne, and K. P. Kording, “Toward an Integration of Deep Learning and Neuroscience,” Front. Comput. Neurosci., vol. 10, p. 94, 2016.

    \item B. A. Richards et al., “A deep learning framework for neuroscience,” Nature Neuroscience, vol. 22, no. 11, pp. 1761–1770, 2019.

    \item A. Saxe, S. Nelli, and C. Summerfield, “If deep learning is the answer, what is the question?,” Nature Reviews Neuroscience, vol. 22, no. 1. Nature Research, pp. 55–67, 2021.

    \item R. M. Cichy and D. Kaiser, “Deep Neural Networks as Scientific Models,” Trends in Cognitive Sciences, vol. 23, no. 4, pp. 305–317, 2019.

    \item L. A. Atherton, D. Dupret, J. R. Mellor, "Memory trace replay: the shaping of memory consolidation by neuromodulation," Trends in Neurosciences, vol. 38, no. 9, pp. 560-570, 2015.

    \item F. Stella, P. Baracskay, J. O’Neill, and J. Csicsvari, “Hippocampal Reactivation of Random Trajectories Resembling Brownian Diffusion,” Neuron, vol. 102, no. 2, p. 450–461.e7, 2019.

    \item J. Kirkpatrick et al., “Overcoming catastrophic forgetting in neural networks,” Proc. Natl. Acad. Sci. U. S. A., vol. 114, no. 13, pp.3521-3526, 2017.

    \item 	Z. Li and D. Hoiem, “Learning without Forgetting,” IEEE Trans. Pattern Anal. Mach. Intell., vol. 40, no. 12, pp. 2935–2947, 2018.

    \item F. Zenke, B. Poole, and S. Ganguli, “Continual learning through synaptic intelligence,” in 34th International Conference on Machine Learning, ICML 2017, vol. 8, pp. 6072–6082, 2017.

    \item N. Y. Masse, G. D. Grant, and D. J. Freedman, “Alleviating catastrophic forgetting using context-dependent gating and synaptic stabilization,” Proc. Natl. Acad. Sci. U. S. A., vol. 115, no. 44, pp. E104657–E104675, 2018.

    \item A. A. Rusu et al., “Progressive Neural Networks,” arXiv preprint arXiv:1606.04671, 2016.

    \item J. Schwarz et al., “Progress \& Compress: A scalable framework for continual learning,” 35th Int. Conf. Mach. Learn. ICML 2018, vol. 10, pp. 7199–7208, 2018.

    \item G. I. Parisi, R. Kemker, J. L. Part, C. Kanan, and S. Wermter, “Continual lifelong learning with neural networks: A review,” Neural Networks, vol. 113, pp. 54–71, 2019.

    \item A. S. Gupta, M. A. A. van der Meer, D. S. Touretzky, and A. D. Redish, “Hippocampal Replay Is Not a Simple Function of Experience,” Neuron, vol. 65, no. 5, pp. 695–705, 2010.

    \item H. F. Ólafsdóttir, C. Barry, A. B. Saleem, D. Hassabis, and H. J. Spiers, “Hippocampal place cells construct reward related sequences through unexplored space,” Elife, vol. 4, 2015.

    \item J. K. Abadchi et al., “Spatiotemporal patterns of neocortical activity around hippocampal sharp-wave ripples,” Elife, vol. 9, 2020.

    \item R. Hadsell, D. Rao, A. A. Rusu, and R. Pascanu, “Embracing Change: Continual Learning in Deep Neural Networks,” Trends in Cognitive Sciences, vol. 24, no. 12, pp. 1028–1040, 2020.

    \item L. J. Lin, “Programming Robots Using Reinforcement Learning and Teaching.,” Proc. Ninth Natl. Conf. Artif. Intell., pp. 781–786, 1991.

    \item J. O’Keefe and J. Dostrovsky, “The hippocampus as a spatial map. Preliminary evidence from unit activity in the freely-moving rat,” Brain Res., vol. 34, no. 1, pp. 171–175, 1971.

    \item J. O’Keefe and L. Nadel, The Hippocampus as a Cognitive Map. Oxford: Oxford University Press, 1978.

    \item G. Buzsaki, Z. Horvath, R. Urioste, J. Hetke, and K. Wise, “High-Frequency Network Oscillation in the Hippocampus,” Science, vol. 256, no. 5059, pp. 1025-1027, 1992.

    \item V. Ego-Stengel and M. A. Wilson, “Disruption of ripple-associated hippocampal activity during rest impairs spatial learning in the rat,” Hippocampus, vol. 20, no. 1, pp. 1-10, 2009.

    \item M. J. Gruber, M. Ritchey, S. F. Wang, M. K. Doss, and C. Ranganath, “Post-learning Hippocampal Dynamics Promote Preferential Retention of Rewarding Events,” Neuron, vol. 89, no. 5, pp. 1110-1120, 2016.

    \item V. P. Murty, A. Tompary, R. A. Adcock, and L. Davachi, “Selectivity in Postencoding Connectivity with High-Level Visual Cortex Is Associated with Reward-Motivated Memory,” J. Neurosci., vol. 37, no. 3, pp. 537-545, 2017.

    \item A. Schapiro, A. E. McDevitt, T. T. Rogers, S. C. Mednick, and K. A. Norman, “Human hippocampal replay during rest prioritizes weakly learned information and predicts memory performance,” Nat. Commun., vol. 9, 3920, 2018.

    \item N. W. Schuck and Y. Niv, “Sequential replay of nonspatial task states in the human hippocampus,” Science, vol. 364, no. 6447, p. eaaw5181, 2019.

    \item G. M. van de Ven, S. Trouche, C. G. McNamara, K. Allen, and D. Dupret, “Hippocampal Offline Reactivation Consolidates Recently Formed Cell Assembly Patterns during Sharp Wave-Ripples,” Neuron, vol., 92, no. 5, pp.968-974, 2016.

    \item L. Roux, B. Hu, R. Eichler, E. Stark, and G. Buzsáki, “Sharp wave ripples during learning stabilize the hippocampal spatial map,” Nat. Neurosci., vol. 20, pp. 845-853, 2017.

    \item D. Marr, “Simple memory: a theory for archicortex,” Phil. Trans. R. Soc. Lond. B, vol. 262, pp. 23–81, 1971.

    \item C. Pavlides and J. Winson, “Influences of hippocampal place cell firing in the awake state on the activity of these cells during subsequent sleep episodes”, J. Neurosci., vol. 9, no. 8, pp. 2907-2918, 1989.

    \item D. Tse, R. F. Langston, M. Kakeyama, I. Bethus, P. A. Spooner, E. R. Wood, M. P. Witter, and R. G. M. Morris, “Schemas and Memory Consolidation,” Science, vol. 316, pp. 76-82, 2007.

    \item J. L. McClelland, “Incorporating Rapid Neocortical Learning of New Schema-Consistent Information Into Complementary Learning Systems Theory,” Journal of Experimental Psychology: General. A, vol. 142, no. 4, p. 1190, 2013.

    \item J. W. Antony and K. A. Paller, “Hippocampal Contributions to Declarative Memory Consolidation During Sleep,” In Hannula, D. E. (Ed.), The Hippocampus from Cells to Systems: Structure, Connectivity, and Functional Contributions to Memory and Flexible Cognition, pp. 245-280, 2017. Cham. Springer International Publishing.

    \item A. C. Schapiro, N. B. Turk-Browne, M. M. Botvinick, and K. A. Norman, “Complementary learning systems within the hippocampus: a neural network modelling approach to reconciling episodic memory with statistical learning,” Phil. Trans. R. Soc. B., vol. 372, no. 1711, 2017.

    \item J. Cichon and W. B. Gan, “Branch-specific dendritic Ca2+ spikes cause persistent synaptic plasticity,” Nature, vol. 520, pp. 180-185, 2015.

    \item Y. Ziv, L. D. Burns, E. D. Cocker, E. O. Hamel, K. K. Ghosh, L. J. Kitsch, A. El Gamal, and M. J. Schnitzer, “Long-term dynamics of CA1 hippocampal place codes,” Nat. Neurosci., vol. 16, no. 3, pp. 264-266, 2013.

    \item G. Rothschild, E. Eban, and L. M. Frank, “A cortical–hippocampal–cortical loop of information processing during memory consolidation,” Nat. Neurosci., vol. 2, no. 2, pp. 251-259, 2017.

    \item G. P. Gava, S. B. McHugh, L. Lefèvre, V. Lopes-dos-Santos., S. Trouche, M. El-Gaby, S. R. Schultz, and D. Dupret, “Integrating new memories into the hippocampal network activity space,” Nat. Neurosci., vol. 24, pp. 326-330, 2021.

    \item J. D. Shin, W. Tang and S. P. Jadhav, “Dynamics of Awake Hippocampal-Prefrontal Replay for Spatial Learning and Memory-Guided Decision Making,” Neuron, vol. 104, no. 6, pp. 1110-1125.e7, 2019.

    \item H. Igata, Y. Ikegaya, and T. Sasaki, “Prioritized experience replays on a hippocampal predictive map for learning,” Proc. Natl. Acad. Sci. U. S. A., vol. 118, no. 1, e2011266118, 2021.

    \item B. Bhattarai, Lee, J. W., and Jung, M. W., “Distinct effects of reward and navigation history on hippocampal forward and reverse replays,” Proc. Natl. Acad. Sci. U. S. A., vol. 117, no. 1, pp. 689-697, 2020.

    \item A. K. Gillespie, D. A. Astudillo Maya, E. L. Denovellis, D. F. Liu, D. B. Kastner, M. E. Coulter, D. K. Roumis, U. T. Eden, and L. M. Frank, “Hippocampal replay reflects specific past experiences rather than a plan for subsequent choice,” bioRxiv, 2021.03.09.434621, 2021.

    \item K. Kaefer, M. Nardin, K. Blahna, and J. Csicsvari, “Replay of Behavioral Sequences in the Medial Prefrontal Cortex during Rule Switching,” Neuron, vol. 106, no. 1, pp. 154-165.e6, 2020.

    \item B. Giri, H. Miyawaki, K. Mizuseki, S. Cheng, and K. Diba, “Hippocampal Reactivation Extends for Several Hours Following Novel Experience,” J. Neurosci., vol. 39, no. 5, pp. 866-875, 2019.

    \item J. P. Grogan, D. Tsivos, L. Smith, B. E. Knight, R. Bogacz, A. Whone, and E. J. Coulthard, “Effects of dopamine on reinforcement learning and consolidation in Parkinson’s disease,” eLife, vol. 6, e26801, 2017.

    \item A. J. Culbreth, J. A. Waltz, M. J. Frank, and J. M. Gold, “Retention of Value Representations Across Time in People With Schizophrenia and Healthy Control Subjects,” Biological Psychiatry: Cognitive Neuroscience and Neuroimaging, vol. 6, no. 4, pp. 420-428, 2021.

    \item J. X. Wang, Z. Kurth-Nelson, D. Kumaran, D. Tirumala, H. Soyer, J. Z. Leibo, D. Hassabis, and M. Botvinick, “Prefrontal cortex as a meta-reinforcement learning system,” Nat. Neurosci., vol. 21, pp- 860–868, 2018.

    \item W. Dabney, Z. Kurth-Nelson, N. Uchida, C. K. Starkweather, D. Hassabis, R. Munos, and M. Botvinick, “A distributional code for value in dopamine-based reinforcement learning,” Nature, vol. 577, pp. 671–675,  2020.

    \item E. L. Roscow, M. W. Jones, and N. F. Lepora, “Behavioural and computational evidence for memory consolidation biased by reward-prediction errors,” bioRxiv, 716290, 2019.

    \item A. Barreto, S. Hou, D. Borsa, D. Silver, and D. Precup, “Fast reinforcement learning with generalized policy updates,” Proc. Natl. Acad. Sci. U. S. A., vol. 117, no. 48, pp. 30079-30087, 2020.

    \item P. Dayan, "Improving Generalization for Temporal Difference Learning: The Successor Representation," Neural Comput., vol. 5, pp. 613–624, 1993.

\end{enumerate}

\end{document}